\documentclass[11pt]{article}

\usepackage[final]{acl}

\usepackage{times}
\usepackage{latexsym}

\usepackage[T1]{fontenc}

\usepackage[utf8]{inputenc}

\usepackage{microtype}

\usepackage{inconsolata}

\usepackage{graphicx}
\usepackage{amsmath}
\usepackage{amsfonts}
\usepackage{makecell}
\usepackage{soul}
\usepackage{adjustbox}
\usepackage{placeins}

\title{When Less is More: Understanding When Token Filtering Helps and Fails in AI-generated Text Detection}

\author{
\textbf{Xiaoyang Han\textsuperscript{1}},
\textbf{Lvxiaowei Xu\textsuperscript{1}},
\textbf{Ming Cai\textsuperscript{2}}\thanks{Corresponding author}
\\
\\
\textsuperscript{1}College of Computer Science and Technology,
\textsuperscript{2}College of Artificial Intelligence
\\
Zhejiang University, Hangzhou, China
\\
\texttt{\{hanxiaoyang, xlxw, cm\}@zju.edu.cn}
}

\begin{document}
\maketitle
\begin{abstract}
The rapid advancement of large language models (LLMs) has made AI-generated text detection increasingly critical. Existing zero-shot detectors assume that more token-level evidence leads to more reliable detection. However, our empirical study challenges this consensus: fewer tokens sometimes work better, retaining only 40\% can yield optimal performance, yet this benefit is not universal. Using the Entropy Gap Score (EGS), we introduce top-$k$ cumulative probability filtering as a diagnostic probe. Across three representative settings, filtering exhibits strikingly different behaviors. We analyze EGS via typical set theory and quantify its dynamics through entropy calibration and distribution analysis. We find that filtering helps for weak source LMs, where low-entropy tokens are harmful, but fails for strong source LMs, where they are not notably harmful. Our work provides the first systematic analysis showing that some tokens are not merely uninformative but systematically harmful due to entropy miscalibration, revealing a two-sided trade-off in token-level detection.

\end{abstract}

\section{Introduction}
AI-generated text detection is becoming increasingly important with the rapid advancement of large language models (LLMs) \citep{wu2025survey}. Existing zero-shot detectors rely on token-level statistical signals and consistently find that detection performance improves with text length \citep{baofast, xutraining, zhu2026exons}.

\begin{figure}[t]
    \centering
    \includegraphics[width=\linewidth]{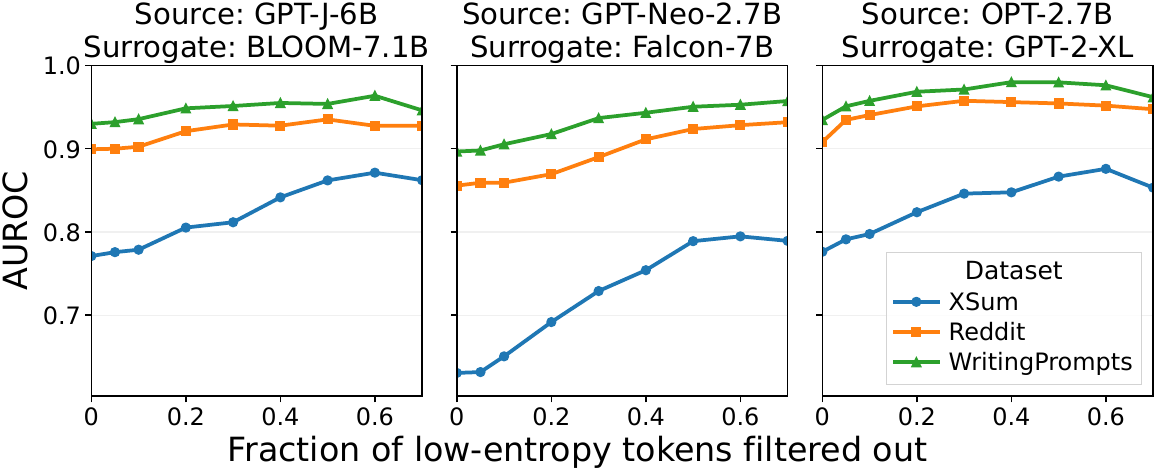}
    \caption{
Filtering out low-entropy tokens improves AUROC across three datasets (XSum, WritingPrompts, Reddit ELI5) under different source-surrogate model pairs. x-axis: filtered-out fraction; y-axis: AUROC.
}
\label{fig:hist-pairs}
\end{figure}

However, our empirical study challenges the consensus that more token-level evidence always leads to better detection. We adopt the Fast-DetectGPT score ~\citep{baofast}, omit normalization for simplicity, and term the resulting score the Entropy Gap Score (EGS). Applying a low-entropy token filtering strategy on XSum~\citep{narayan2018don}, WritingPrompts~\citep{fan2018hierarchical}, and Reddit ELI5~\citep{fan2019eli5}, we find that performance generally improves with more token removal (Figure~\ref{fig:hist-pairs}). Retaining only 40\% of tokens yields an improvement of over 15\%, suggesting that detectors can sometimes benefit from shorter text.

To assess the generalizability of token filtering for detection improvement, we introduce top‑$k$ cumulative probability filtering as a diagnostic probe for token-level analysis. 
This probe captures probability concentration and is more robust than directly filtering by entropy.

By combining source and surrogate models, we construct three representative detection settings: weak‑source black‑box, weak‑source white‑box, and strong‑source black‑box. Results show a stable pattern: filtering consistently improves performance in the weak‑source black‑box setting, but behaves very differently in the others. This finding motivates us to explore the underlying mechanisms.

We first examine EGS, which measures the gap between a token's self-information and its expected self-information (entropy). Using typical set theory~\citep{cover2006elements}, we reinterpret EGS as a signed typicality deviation score. This perspective reveals that human text is closer to the typical set, whereas AI-generated text tends to deviate from it, producing a detectable EGS difference.

To quantify EGS's dynamics, we introduce entropy calibration \citep{caoentropy} and measure it by analyzing the coupling strength between self-information $L$ and entropy $E$ across different entropy intervals. We find that weak source LMs exhibit significant $L$--$E$ decoupling, causing abnormally high EGS in low-entropy regions, which harms detection.

Based on the above analysis, we view top‑$k$ filtering as a two-sided trade-off: removing low-entropy tokens separates AI from human distributions (beneficial), but filtering also reduces token count and increases EGS variance (harmful). Whether the net gain is positive depends on the source-side entropy calibration.

Our contributions are summarized as follows:

\begin{itemize}
  \item We empirically show that removing tokens can substantially improve detection, challenging the consensus that more token evidence is always better. This provides the first systematic analysis that some tokens are not merely uninformative but systematically harmful.
  \item We introduce top-$k$ filtering as a diagnostic probe to analyze token-level signals, revealing distinct behaviors across three detection settings and showing that low-entropy tokens can be harmful in weak source LMs due to entropy miscalibration.
  \item We reinterpret EGS via typical set theory, quantify its dynamics using entropy calibration and distribution analysis, and uncover a two-sided trade-off (shift vs. variance) that governs filtering success, explaining the distinct behaviors across the three settings.
\end{itemize}

\section{Related Work}
\noindent\textbf{Training-free detectors} for AI-generated text, also known as zero-shot detectors, generally exhibit better generalization ability than training-based methods and typically rely on token-level statistical features. Early methods use likelihood-based token statistics, such as log likelihood~\citep{hashimoto2019unifying}, entropy~\citep{gehrmann2019gltr}, and log-rank~\citep{su2023detectllm}. 

DetectGPT~\citep{mitchell2023detectgpt} and Fast-DetectGPT~\citep{baofast} show that human-written and AI-generated text differ in the curvature of the log-probability function. They measure this difference using detection scores that are closely related to the Entropy Gap Score (EGS).
Recently, researchers have begun to model token-level statistics as time series and analyze them in either time domain (e.g., Lastde++~\citep{xutraining}) or frequency domain (e.g., SpecDetect++~\citep{luo2026specdetect}).
In general, these methods aggregate statistical signals across all tokens to compute a detection score under the assumption that more token evidence leads to more reliable detection. However, we find that removing entropy-miscalibrated tokens can improve detection performance.


\noindent\textbf{Token importance imbalance} is suggested by recent studies, with only a small subset of important tokens playing a dominant role.  Exons-Detect~\citep{zhu2026exons} distinguishes informative exonic tokens from less useful intronic tokens, highlighting their unequal contributions to detection. 
\citet{mohammadi2025explainability} identify the most influential tokens in AI-generated text and reduce detectability by replacing them. 
\citet{fangwrong} show that standard perplexity is ineffective because it treats all tokens equally, although only a small subset of key tokens determines long-context performance. 
H2O~\citep{zhang2023h2o} finds that a few heavy-hitter tokens dominate attention scores and are critical for generation quality during KV cache eviction. 
Prior work mainly focuses on identifying informative tokens, while our work further reveals that some tokens can be systematically harmful.

\noindent\textbf{Typical set} consists of sequences whose probabilities are close to the entropy of source distribution, and is a fundamental concept in information theory \citep{cover2006elements}. 
Beyond theory, it has been applied to language modeling tasks. \citet{hashimoto2019unifying} use human ratings of sentence typicality and model probabilities to detect machine-generated text.
\citet{meister2022typical,meister2023locally} propose typical sampling, which decodes tokens from the typical set to generate more natural text. \citet{nalisnick2019detecting} utilize the typical set for out-of-distribution detection by determining whether test samples belong to the typical set. 
Our work further exploits the typical set to investigate the underlying mechanism of AI-generated text detection.



\section{Preliminaries and Filtering Probe}
\subsection{AI-Generated Text Detection}
\noindent\textbf{Task.} 
We define AI-generated text detection as a binary classification task that distinguishes human-written text (HWT) from AI-generated text (AIGT).

\noindent\textbf{Source and Surrogate Models.}
Given an input text $x = (x_1, \ldots, x_N)$, we use a surrogate model $M$ (or scoring LM) to extract token-level features. The source model is the one that originally generates the text.
When the surrogate model is the source model itself, it is the white-box setting; otherwise, it is the black-box setting~\citep{baofast}.

\noindent\textbf{Weak and strong LMs.}
To analyze the impact of source and surrogate models, we categorize LMs as weak or strong. Weak LMs are smaller or older open-source models (e.g., Falcon-7B, GPT-2-XL). Strong LMs are newer, larger models (e.g., GPT-4o, GPT-4-Turbo).

\subsection{Formulation of Entropy Gap Score}
Given an N-token input $x$ and a surrogate model $M$, we define the self-information $L_i$ and expected self-information (entropy) $E_i$ for token $x_i$ as:

\begin{equation}
L_i = -\log p_M(x_i \mid x_{<i})
\end{equation}
\begin{equation}
\begin{aligned}
E_i
&= \mathbb{E}_{v\sim p_M(\cdot \mid x_{<i})}
\left[-\log p_M(v \mid x_{<i})\right] \\
&= -\sum_{v\in\mathcal{V}}
p_M(v\mid x_{<i})
\log p_M(v\mid x_{<i}),
\end{aligned}
\end{equation}
where $\mathcal{V}$ denotes the vocabulary. 

We compute the Entropy Gap Score (EGS) as the mean of ($L_i$ - $E_i$) over all tokens:
\begin{equation}
\mathrm{EGS}(x)
= \frac{1}{N}\sum_{i=1}^{N}(L_i - E_i).
\end{equation}
EGS subtracts entropy from the observed self-information, thus calibrating token likelihood against the surrogate model's uncertainty. Compared to Fast-DetectGPT~\citep{baofast}, EGS simplifies the score by omitting normalization. The detailed derivation is in Appendix~\ref{app:derivation}. EGS is the primary detection score used in this paper.

\subsection{Token Filtering Probe}
We introduce top-$k$ filtering as a probe to analyze token-level detection signals. Instead of treating all tokens equally, it selectively removes low-entropy tokens based on probability concentration, enabling us to study how different token subsets affect detection behavior. This probe is not merely for performance optimization, but serves as a tool for analyzing entropy calibration, $L$--$E$ coupling in EGS, and harmful token signals in AIGT detection.

Concretely, inspired by top-$p$ sampling~\citep{holtzmancurious}, we measure the probability concentration at each position by the cumulative probability mass of the top-$k$ vocabulary tokens:
\begin{equation}
m_i^{(k)}
= \sum_{j=1}^{k} p_M(v_j \mid x_{<i}).
\end{equation}
Let $\theta\in[0,1)$ be the filtering ratio and
$r_\theta=\lfloor \theta N\rfloor$. We define $U_\theta$ as the set of
$r_\theta$ token positions with the largest $m_i^{(k)}$ values:
\begin{equation}
U_\theta
= \operatorname{Top}_{r_\theta}
\left(\{m_i^{(k)}\}_{i=1}^{N}\right).
\end{equation}
Unless otherwise specified, we set $k=10$ in all experiments.
Appendix~\ref{app:k_sensitivity} shows that the filtering pattern remains
stable when $k$ changes. The filtered Entropy Gap Score is
\begin{equation}
\mathrm{EGS}_{\theta}(x)
=
\frac{1}{N-r_\theta}
\sum_{i \notin U_\theta} (L_i - E_i).
\end{equation}
When $\theta=0$, this reduces to the unfiltered EGS.

\section{Experimental Setup}

\subsection{Datasets}
We use three datasets covering different domains: XSum~\citep{narayan2018don} (BBC News documents), WritingPrompts~\citep{fan2018hierarchical} (for story generation), and Reddit ELI5 \citep{fan2019eli5} (Q\&A data restricted to the topics of biology, physics, chemistry, economics, law, and technology). For each dataset, we randomly sample 150 human-written examples and take the first 30 tokens of each as the prompt to the source model.
Appendix~\ref{app:sample_robustness} further shows that this sample size is sufficient.

\subsection{Source and Surrogate Models}

\noindent\textbf{Source models.}
For comprehensive analysis, we use diverse source models to generate text: six weak LMs (GPT-2-XL, OPT-2.7B, GPT-Neo-2.7B, GPT-J-6B, Falcon-7B, and BLOOM-7.1B) and five strong LMs (GPT-4-Turbo, GPT-4o, Claude-4, DeepSeek-V4, Gemini-2.5). 
For the controlled validation in Section~\ref{sec:controlled_source}, we additionally include Falcon-7B-Instruct as a weak source LM.

\noindent\textbf{Surrogate models.}
We use the same six weak LMs as the main open-source surrogate models to compute the Entropy Gap Score.
For the cross-surrogate analysis, we additionally include Falcon-40B and Qwen2.5-3B.
For all source and surrogate models, see Appendix~\ref{app:implementation-details}
for details and citations.

\begin{table*}[t]
  \centering
  \small
  \setlength{\tabcolsep}{3.8pt}
  \renewcommand{\arraystretch}{1.18}
  \begin{tabular}{l*{12}{c}}
    \hline
    \textbf{Methods}
    & \multicolumn{2}{c}{\textbf{Falcon-7B}}
    & \multicolumn{2}{c}{\textbf{GPT-J}}
    & \multicolumn{2}{c}{\textbf{OPT-2.7B}}
    & \multicolumn{2}{c}{\textbf{GPT-Neo-2.7B}}
    & \multicolumn{2}{c}{\textbf{BLOOM-7.1B}}
    & \multicolumn{2}{c}{\textbf{Avg.}} \\
    \cline{2-13}

    & \textbf{Self} & \textbf{GPT2}
    & \textbf{Self} & \textbf{GPT2}
    & \textbf{Self} & \textbf{GPT2}
    & \textbf{Self} & \textbf{GPT2}
    & \textbf{Self} & \textbf{GPT2}
    & \textbf{Self} & \textbf{GPT2} \\
    \hline

    Log-Likelihood    & 0.7940 & 0.6544 & 0.8708 & 0.7570 & 0.8894 & 0.7243 & 0.9099 & 0.7539 & 0.8994 & 0.6761 & 0.8727 & 0.7131 \\
    Entropy           & 0.4495 & 0.4916 & 0.5001 & 0.5154 & 0.4969 & 0.4766 & 0.5224 & 0.4764 & 0.4203 & 0.4347 & 0.4778 & 0.4789 \\
    LogRank           & 0.8327 & 0.7101 & 0.9013 & 0.8035 & 0.9147 & 0.7835 & 0.9378 & 0.8024 & 0.9355 & 0.7484 & 0.9044 & 0.7696 \\
    DetectGPT         & 0.8498 & 0.6258 & 0.9315 & 0.7323 & 0.9360 & 0.7655 & 0.9693 & 0.7934 & 0.9714 & 0.6707 & 0.9316 & 0.7175 \\
    Fast-DetectGPT    & 0.9751 & 0.7632 & 0.9864 & 0.8741 & 0.9857 & 0.8730 & \underline{0.9949} & 0.8990 & 0.9956 & 0.8507 & 0.9875 & 0.8520 \\
    Lastde++          & \textbf{0.9889} & \underline{0.8122} & \textbf{0.9940} & \underline{0.9217} & \textbf{0.9937} & \underline{0.9204} & \textbf{0.9990} & \underline{0.9443} & \textbf{0.9995} & \underline{0.8953} & \textbf{0.9950} & \underline{0.8988} \\
    SpecDetect++      & \underline{0.9839} & 0.7818 & \underline{0.9894} & 0.9026 & \underline{0.9898} & 0.8984 & 0.9945 & 0.9267 & \underline{0.9961} & 0.8710 & \underline{0.9907} & 0.8761 \\
    EGS (0\%)         & 0.9759 & 0.7647 & 0.9845 & 0.8746 & 0.9848 & 0.8729 & 0.9935 & 0.8994 & 0.9958 & 0.8511 & 0.9869 & 0.8525 \\
    EGS (50\%)        & 0.9802 & \textbf{0.8945} & 0.9857 & \textbf{0.9372} & 0.9819 & \textbf{0.9365} & 0.9942 & \textbf{0.9572} & 0.9937 & \textbf{0.9413} & 0.9872 & \textbf{0.9333} \\

    \hline
    
    EGS (opt)     & 0.9807 & 0.8948 & 0.9888 & 0.9378 & 0.9861 & 0.9371 & 0.9953 & 0.9579 & 0.9971 & 0.9438 & 0.9896 & 0.9343 \\
    
    optimal~$\theta$ & 0.57 & 0.53 & 0.47 & 0.53 & 0.30 & 0.53 & 0.37 & 0.53 & 0.20 & 0.47 & 0.38 & 0.52 \\
    \hline
  \end{tabular}

\caption{
Top-$k$ filtering performance on five weak source LMs across three datasets (XSum, WritingPrompts, Reddit). ``Self'': white-box setting; ``GPT2'': black-box setting (surrogate: GPT-2-XL). 
EGS ($\theta$): AUROC at fixed ratio $\theta$.
Average AUROC over three datasets is reported, with best non-oracle performance highlighted and second-best underlined. 
For oracle analysis, EGS (opt): best AUROC over filtering ratios; ``optimal $\theta$'': average optimal filtering ratio across datasets.
}

  
  \label{tab:weak_results}
\end{table*}

\subsection{Baselines}
We evaluate top-$k$ filtering against representative training-free detectors. For token-statistics-based detectors, we use Log-Likelihood~\citep{hashimoto2019unifying}, Entropy~\citep{gehrmann2019gltr}, and LogRank~\citep{su2023detectllm}. For perturbation-based detectors, we use DetectGPT~\citep{mitchell2023detectgpt} and Fast-DetectGPT~\citep{baofast}. For time-series-based detectors, we use Lastde++~\citep{xutraining} and SpecDetect++~\citep{luo2026specdetect}.
The unfiltered EGS serves as an additional baseline.

\subsection{Metrics}
We evaluate detection performance using AUROC~\citep{mitchell2023detectgpt, baofast}, which does not require a fixed decision threshold. 

Since top-$k$ filtering serves as a diagnostic probe rather than a deployment method, we report the oracle best achievable AUROC across filtering ratios and the corresponding \textbf{optimal filtering ratio}.

\section{Empirical Analysis of Token Filtering}
\label{sec:experiment}
\subsection{Filtering Helps in Weak Source LMs}
\label{sec:weak_source}

We evaluate top-$k$ filtering on weak source LMs under both black-box and white-box settings.

\noindent\textbf{Black-box setting.} 
We evaluate three top-$k$ filtering variants, denoted as EGS (0\%), EGS (50\%), and EGS (opt) according to their filtering ratios. We compare them against seven baselines across five weak source LMs with a fixed surrogate model (GPT-2-XL) on three datasets. Average AUROC is reported. See Appendix~\ref{app:weak_details} for detailed results.


As shown in Table~\ref{tab:weak_results}, EGS (50\%) ranks first among non-oracle methods in AUROC (averaging 0.9333). It exceeds the second-best model Lastde++ (0.8988) by 3.45 points, with its best gain (8.23 points) on Falcon-7B (0.8122$\to$0.8945).
EGS (opt) reaches a slightly higher average AUROC of 0.9343, close to EGS (50\%). 

Notably, the optimal filtering ratio averages 0.52, meaning retaining 48\% of tokens yields the best performance. Consistently, held-out ratio selection also chooses 0.5, as detailed in Appendix~\ref{app:ratio_selection}.
Compared to the unfiltered EGS (0\%), EGS (opt) achieves an average improvement of 8.18 points (0.8525$\to$0.9343), with its best gain (13.01 points) on Falcon-7B (0.7647$\to$0.8948). 
This indicates that top-$k$ filtering achieves substantial performance gains with fewer tokens.

\noindent\textbf{White-box setting.} EGS (opt) and EGS (50\%) achieve comparable performance across all source LMs. Despite the already-high detection performance in the white-box setting, EGS (opt) improves over EGS (0\%) by 0.27 points (0.9869$\to$0.9896) at an optimal ratio of 0.38, with its largest gain (0.48 points) on Falcon-7B (0.9759$\to$0.9807).

In general, token filtering benefits weak source LMs in both black-box and white-box settings. We explore the underlying mechanism in Section~\ref{sec:helps_and_fails}.

\subsection{Filtering Fails in Strong Source LMs}
\label{sec:strong_source}
We examine whether top-$k$ filtering generalizes to stronger source LMs by evaluating three top-$k$ filtering variants against four baselines across two strong source LMs (GPT-4-Turbo, GPT-4o) with a fixed surrogate (GPT-2-XL) on three datasets, and report average AUROC.

\begin{table}[t]
  \centering
  \small
  \setlength{\tabcolsep}{3pt}
  \resizebox{\columnwidth}{!}{
  \begin{tabular}{lcccc}
    \hline
    \textbf{Method}
    & \textbf{XSum} & \textbf{WritingPrompts} & \textbf{Reddit} & \textbf{Avg.} \\
    \hline
    \multicolumn{5}{c}{\textbf{GPT-4-Turbo}} \\
    \hline
    DetectGPT           & 0.5392 & 0.7517 & 0.7824 & 0.6911 \\
    Fast-DetectGPT      & 0.7819 & 0.7524 & 0.9238 & 0.8194 \\
    Lastde++            & 0.7698 & 0.7434 & 0.9194 & 0.8109 \\
    SpecDetect++        & 0.7297 & 0.6986 & 0.8176 & 0.7486 \\
    EGS (0\%)           & \textbf{0.7840} & \textbf{0.7534} & \textbf{0.9271} & \textbf{0.8215} \\
    EGS (50\%)          & 0.7596 & 0.7288 & 0.8873 & 0.7919 \\
    \hline
    EGS (opt)
        & 0.7934
        & 0.7625
        & 0.9271
        & 0.8277 \\
    optimal~$\theta$    & 0.10  & 0.10   & 0   & 0.07      \\
    \hline
    \multicolumn{5}{c}{\textbf{GPT-4o}} \\
    \hline
    DetectGPT           & 0.5478 & 0.7876 & 0.8991 & 0.7448  \\
    Fast-DetectGPT      & \textbf{0.8553} & \textbf{0.8619} & \textbf{0.9696} & \textbf{0.8956} \\
    Lastde++            & 0.8267 & 0.8504 & 0.9609 & 0.8793  \\
    SpecDetect++        & 0.7226 & 0.7992 & 0.9168 & 0.8129  \\
    EGS (0\%)           & 0.8232 & 0.8378 & 0.9685 & 0.8765 \\
    EGS (50\%)          & 0.7879 & 0.7997 & 0.9329 & 0.8402 \\
    \hline
    EGS (opt)           & 0.8273 & 0.8424 & 0.9685 & 0.8794 \\
    optimal~$\theta$    & 0.20    & 0.10    & 0      & 0.10    \\
    \hline
  \end{tabular}
  }
  \caption{Detection results on two strong source LMs (GPT-4-Turbo, GPT-4o) with a fixed surrogate (GPT-2-XL) on three datasets. Average AUROC is reported. ``optimal $\theta$'' denotes the best filtering ratio.}
  \label{tab:strong_results}
\end{table}

As shown in Table~\ref{tab:strong_results}, interestingly, the improvement disappears for strong source LMs, where filtering consistently degrades or becomes ineffective. Specifically, compared with EGS (0\%), EGS (50\%) drops the average AUROC by 2.96 points on GPT-4-Turbo (0.8215$\to$0.7919) and by 3.63 points on GPT-4o (0.8765$\to$0.8402).

This ineffectiveness is also reflected in the optimal filtering ratio, which remains close to zero for strong source LMs. Specifically, the average optimal ratio is 0.07 on GPT-4-Turbo and 0.10 on GPT-4o, indicating that filtering provides little benefit for strong source LMs.

These results contrast sharply with the weak-source settings, showing that the effectiveness of filtering strategies depends on the source LM and is not universally beneficial. We investigate when and why filtering strategies fail in Section~\ref{sec:helps_and_fails}.
We further validate this strong-source pattern on additional source LMs in Section~\ref{sec:controlled_source}.

\begin{figure}[t]
    \centering
    \includegraphics[width=\linewidth]{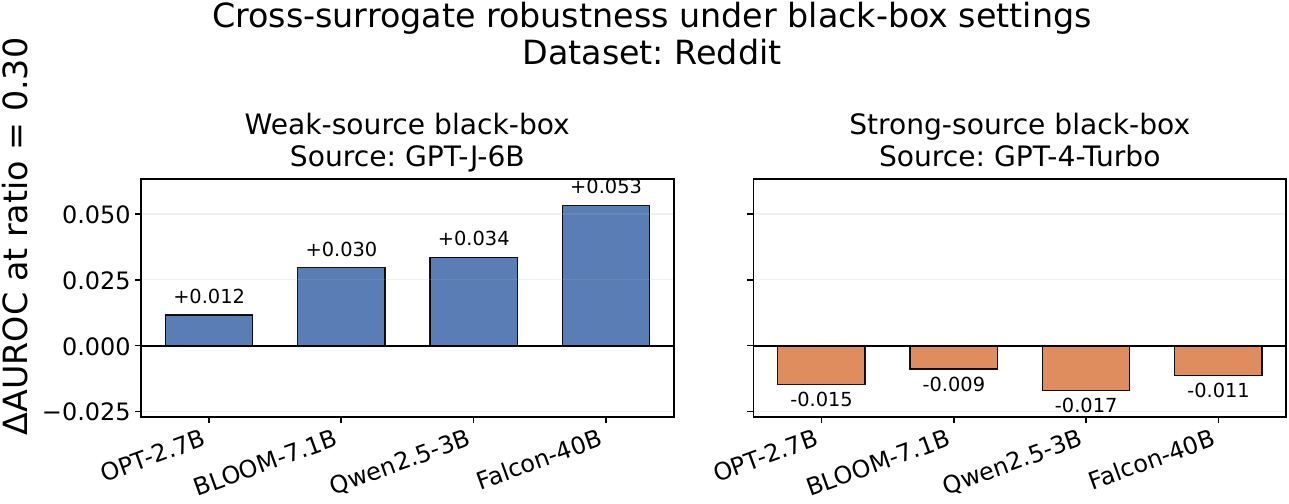}
    \caption{
Cross-surrogate effect of top-$k$ filtering on two source LMs (weak: GPT-J-6B; strong: GPT-4-Turbo) with four surrogates (OPT-2.7B, BLOOM-7.1B, Qwen2.5-3B, Falcon-40B) on Reddit.
Bars show AUROC change after filtering (relative to unfiltered EGS (0\%)) at a fixed 30\% ratio. Left: weak source; right: strong source.
}
\label{fig:cross_surrogate}
\end{figure}

\begin{figure}[t]
    \centering
    \includegraphics[width=\linewidth]{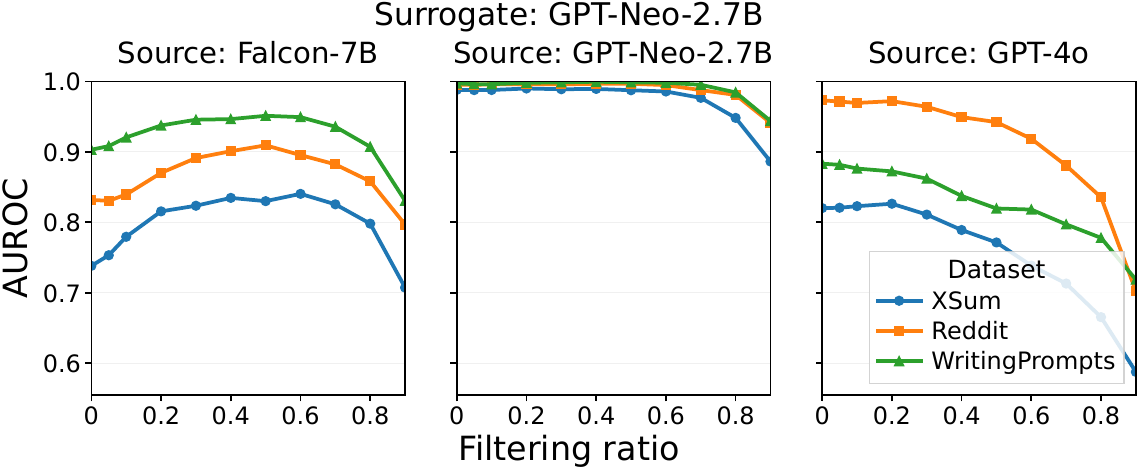}
    \caption{
Top-$k$ filtering performance under varying filtering ratios on three datasets with a fixed surrogate (GPT-Neo-2.7B) and three source LMs (left to right): Falcon-7B (weak-source black-box), GPT-Neo-2.7B (weak-source white-box), GPT-4o (strong-source black-box). AUROC is reported.}

\label{fig:filtering_ratio}
\end{figure}

\subsection{Cross-Surrogate Consistency}
\label{sec:cross_surrogate}
In previous sections (Section~\ref{sec:weak_source} and Section~\ref{sec:strong_source}), we use a fixed surrogate model (GPT-2-XL). To rule out that this specific choice explains the weak--strong source LM difference, we evaluate top-$k$ filtering on two source LMs (weak: GPT-J-6B; strong: GPT-4-Turbo) with four surrogate models of different scales (OPT-2.7B, BLOOM-7.1B, Qwen2.5-3B, Falcon-40B) on Reddit. We report the AUROC change after filtering (relative to unfiltered EGS (0\%)) at a fixed 30\% ratio.

As shown in Figure~\ref{fig:cross_surrogate}, the weak-source setting consistently yields positive gains across surrogates (+0.012 to +0.053), while the strong-source setting shows consistent degradation (-0.017 to -0.009). Although the $\Delta$AUROC magnitude varies, its direction remains stable across surrogates.
This indicates that different surrogates affect the strength, not the sign, of the filtering effect.

Moreover, to rule out surrogate-specific bias, we examine the removed-token sets across different surrogates and tokenizers.
Appendix~\ref{app:cross_surrogate_filtering} shows that the selected regions exhibit substantially higher overlap than random selection under both shared- and different-tokenizer settings. 
Therefore, the filtering behavior difference mainly reflects source-side features motivating our entropy calibration analysis in Section~\ref{sec:entropy_calibration}.

\subsection{Filtering Ratio Analysis}
\label{sec:filtering_ratio}

\begin{figure*}[t]
    \centering
    \includegraphics[width=0.90\linewidth]{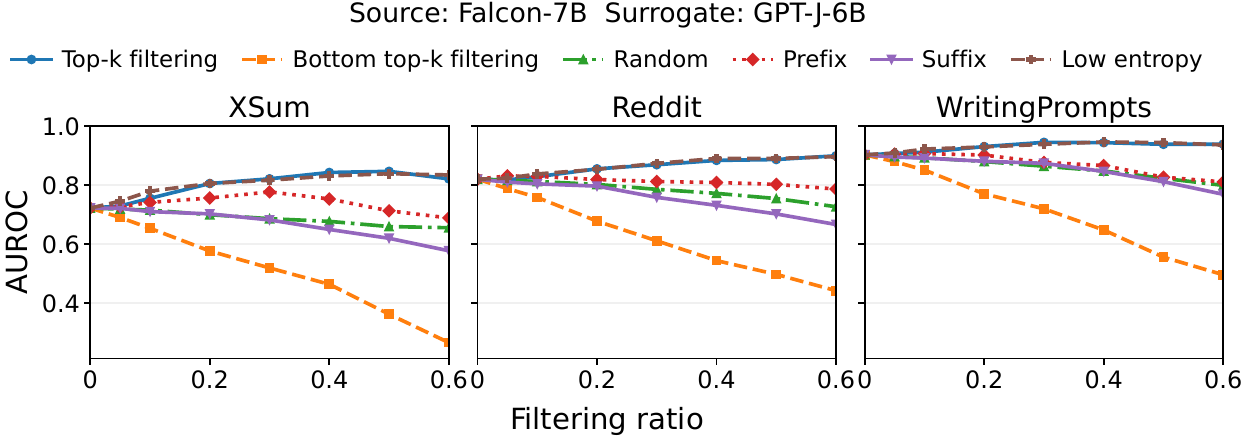}
    \caption{
Comparison of six filtering strategies under weak-source black-box (Falcon-7B/GPT-J-6B) on three datasets. x-axis: filtering ratio; y-axis: AUROC. "Prefix/Suffix" = remove beginning/end of the original sequence.
}
\label{fig:entropy-align}
\end{figure*}

To investigate top-$k$ filtering across filtering ratios, we construct three setups with a fixed surrogate (GPT-Neo-2.7B) and three source LMs: Falcon-7B (weak-source black-box), GPT-Neo-2.7B (weak-source white-box), and GPT-4o (strong-source black-box). These are denoted as \textbf{Weak-Black}, \textbf{Weak-White} and \textbf{Strong-Black}. We evaluate filtering on three datasets and report AUROC for ratios from 0 to 1 in steps of 0.1.

Figure~\ref{fig:filtering_ratio} presents three distinct patterns of AUROC across the three experimental setups.
For the \textbf{Weak-Black} setup, AUROC initially increases with the filtering ratio, reaches its peak around 0.5, and then gradually declines.
Unlike the inverted U-shaped pattern observed in \textbf{Weak-Black}, the \textbf{Weak-White} setup exhibits a plateau up to a filtering ratio of 0.6, followed by a monotonic decrease.
In contrast, the \textbf{Strong-Black} setup shows no initial plateau. Performance begins to decline at a low filtering ratio (around 0.2) and continues to drop.

We analyze and explain the mechanisms underlying these three patterns in Section~\ref{sec:helps_and_fails}.

\subsection{Comparison with Alternative Filtering Strategies}
\label{sec:alternative_strategies}

To verify that the gain depends on \textit{which} tokens are removed, not just \textit{how many}, we compared top-$k$ filtering with five alternatives (low-entropy, random, bottom top-$k$, prefix, suffix) under a representative weak-source black-box setting (source: Falcon-7B; surrogate: GPT-J-6B) on three datasets (XSum, Reddit and WritingPrompts). Bottom top-$k$ filtering removes tokens with the lowest probability mass, while prefix and suffix filtering remove tokens from the beginning or end of the sequence, respectively.

As shown in Figure~\ref{fig:entropy-align}, top-$k$ filtering performs best, followed by low-entropy filtering with a similar but weaker trend.
In contrast, random, prefix, and suffix filtering degrade performance, indicating that arbitrary filtering does not yield gains; only entropy-aligned filtering provides benefits.
Notably, bottom top-$k$ filtering performs worst, as it removes the most discriminative tokens.

\begin{figure}[t]
    \centering
    \includegraphics[width=0.95\linewidth]{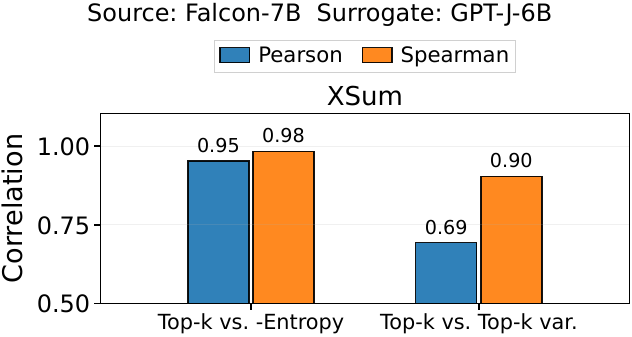}
    \caption{
Correlation between top-$k$ probability mass and related concentration features, including negative entropy and top-$k$ probability variance. Pearson and Spearman correlations are shown.
}
\label{fig:feature-corr}
\end{figure}

As a sanity check, Figure~\ref{fig:feature-corr} shows that top-$k$ probability mass is highly correlated with negative entropy (Pearson $=0.95$) and moderately correlated with top-$k$ probability variance (Pearson $=0.69$), confirming that top-$k$ filtering mainly targets tokens that are low-entropy and high-concentration.

\subsection{Robustness Under Rewriting Attacks}
\label{sec:attack_robustness}

To assess whether our conclusions hold under more realistic scenarios, we evaluate robustness under two rewriting attacks: paraphrasing and back-translation. 
Fixed-ratio filtering consistently improves unfiltered EGS under both rewriting transformations and outperforms Lastde++ in all six settings. Detailed results are provided in Appendix~\ref{app:rewriting}.








\section{Understanding Token Filtering}
\label{sec:mechanism}
\subsection{Entropy Gap and Typicality}
\label{sec:typicality}
Top-$k$ filtering is based on the Entropy Gap Score (EGS); thus, explaining its behavior requires understanding EGS's mechanism.

EGS measures the difference between a token's self-information $L_i$ and its expected self-information (entropy) $E_i$ from the surrogate model's perspective. Prior work attributes this detection signal to probability and conditional probability curvature~\citep{mitchell2023detectgpt,baofast}. We reinterpret EGS from a typicality perspective as a signed typicality deviation score. For a token typical under the surrogate distribution, its self-information should approximate the model's entropy expectation. Human-written text shows a smaller EGS discrepancy~\citep{meister2022typical,meister2023locally}, while AI-generated text often deviates from the typical set with a systematic EGS shift, providing a discriminative signal for detection.

However, typicality captures only the overall EGS trend. Fully explaining top-$k$ filtering requires examining the quantitative relationships within EGS: entropy calibration, the EGS distribution, and how EGS changes under different conditions.

\begin{figure*}[t]
    \centering
     \includegraphics[width=\linewidth]{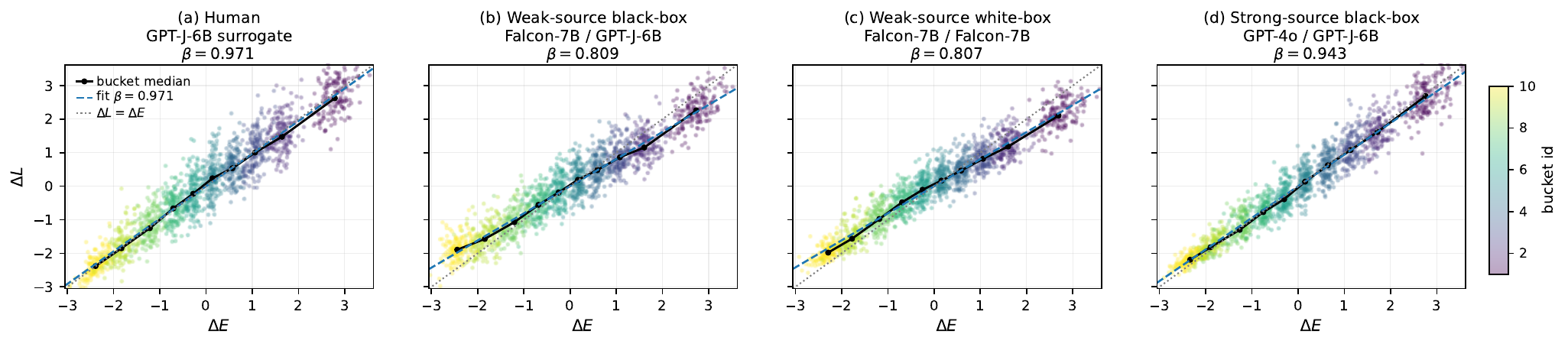}
    \caption{
Linear regression of centered self-information $L$ on centered entropy $E$ across ten entropy buckets (XSum).
$\beta$: regression coefficient.
Panels: Human, Weak-black, Weak-white, Strong-black (definitions in Section~\ref{sec:entropy_calibration}).
}
\label{fig:le-coupling}
\end{figure*}

\subsection{Entropy Calibration and L–E Coupling}
\label{sec:entropy_calibration}

We define entropy calibration \citep{caoentropy} as how closely the observed self-information $L$ follows its expected self-information $E$. In a well-calibrated model, $L$ and $E$ co-vary strongly, keeping the gap ($L$ - $E$) stable across tokens.

To quantify calibration, we analyze the coupling between $L$ and $E$ across entropy levels.
For each text, tokens are sorted by entropy into ten buckets (from high to low entropy) and we compute the mean self-information $\bar{L}_b$ and mean entropy $\bar{E}_b$ for each bucket b.
To remove text‑specific offsets, we center these values by subtracting the corresponding text mean: 
$\Delta L_b=\bar{L}_b-\bar{L}_{\mathrm{text}}$ and
$\Delta E_b=\bar{E}_b-\bar{E}_{\mathrm{text}}$.

We fit a linear regression over all centered bucket means. The slope $\beta$ quantifies the coupling strength: $\beta \approx 1$ indicates strong coupling (good calibration), while $\beta < 1$ indicates miscalibration, where EGS varies more across entropy levels.

As shown in Figure~\ref{fig:le-coupling}, human-written text exhibits the strongest $L$--$E$ coupling ($\beta = 0.971$), consistent with prior work \citep{meister2022typical,meister2023locally}. For AI-generated text, we analyze three settings on XSum: Weak-black (Falcon-7B/GPT-J-6B), Weak-white (Falcon-7B/Falcon-7B), and Strong-black (GPT-4o/GPT-J-6B). These abbreviations are used throughout the following analyses.

The Strong-black setting also shows a strong coupling pattern ($\beta = 0.943$). In contrast, the weak-source setting exhibits weaker coupling: $\beta = 0.809$ (Weak-black) and $\beta = 0.807$ (Weak-white). This indicates that EGS varies more across entropy buckets for weak source LMs. 
Appendix~\ref{summary_all_pairs} confirms this trend across all source-surrogate pairs and datasets.

These observations suggest stronger entropy miscalibration in the weak-source settings than in the strong-source settings.

\subsection{Why Filtering Helps and Fails}
\label{sec:helps_and_fails}

\begin{figure}[t]
\includegraphics[width=0.65 \linewidth]{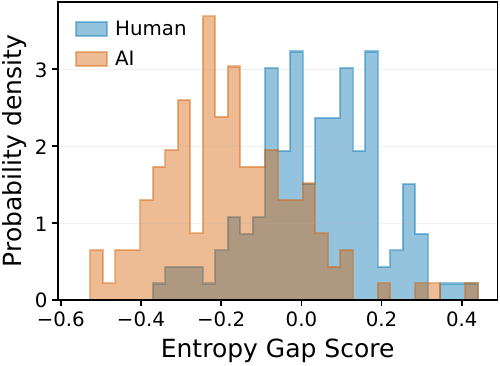}
\centering
\caption{
EGS distributions of human-written and AI-generated text in a weak-source, black-box setting.
}
\label{fig:score-hist}
\end{figure}

\noindent\textbf{EGS distribution.} Zero-shot detection performance depends on how well the detection score (e.g., EGS, Lastde++) separates human-written and LLM-generated texts~\citep{mitchell2023detectgpt,xutraining}. As shown in Figure~\ref{fig:score-hist}, AI-side EGS typically lies left of human distribution. Due to opposite sign, DetectGPT and Fast-DetectGPT show the reverse (AI on the right). 
Therefore, when filtering shifts the AI-side distribution further left, the overlap between the two distributions decreases, improving class separability.

\noindent\textbf{EGS by entropy region.} 
To examine EGS variation with entropy, tokens from human text and three settings  on XSum are split into high- and low-entropy groups (top/bottom 20\% and 50\%). Mean EGS and between-group difference  are computed.

Table~\ref{tab:direction_analysis} reveals four observations: 1) Human and Strong-black settings show much smaller EGS differences than Weak-black and Weak-white, reflecting differences in entropy calibration across models; 2) In the Weak-black setting, low-entropy tokens exhibit significantly higher EGS, entering the human EGS region and thus having a harmful effect; 3) In the Weak-white setting, low-entropy EGS partially overlaps with the human EGS region, still interfering with detection; 4) In the Strong-black setting, the EGS difference between low- and high-entropy tokens is small, so low-entropy tokens are not notably harmful.


\begin{figure}[t]
\includegraphics[width=0.8      \linewidth]{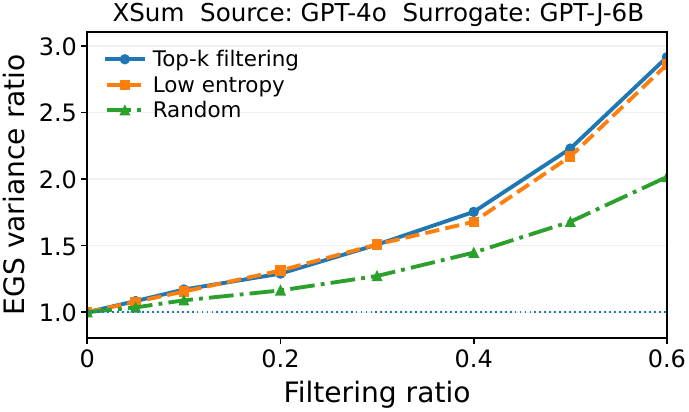}
\centering
\caption{
Comparison of three filtering strategies on XSum under strong-source black-box setting (GPT-4o/GPT-J-6B). x-axis: filtering ratio; y-axis: filtered-to-original EGS variance ratio.
}
\label{fig:variance_ratio}
\end{figure}

\begin{figure}[t]
\includegraphics[width=\linewidth]{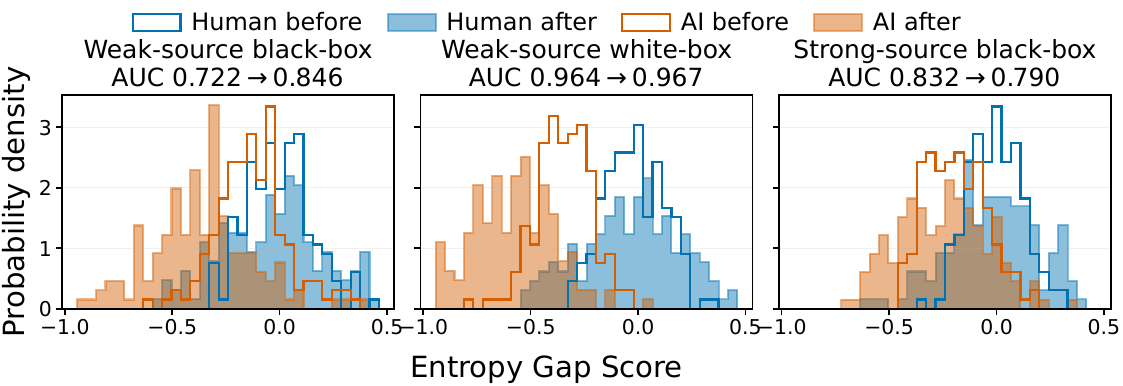}
\caption{
EGS distributions before/after top-$k$ filtering on XSum (50\% ratio). Panels: Weak-black, Weak-white, Strong-black. Histograms are density-normalized.
}
\label{fig:filtering_histogram}
\end{figure}

\begin{table}
\centering
\small
\setlength{\tabcolsep}{1.8pt}

\begin{tabular}{lcccccc}
\hline
Setting & Low-20 & High-20 & Diff & Low-50 & High-50 & Diff \\
\hline
Weak-black   & 0.29  & -0.58 & 0.87 & 0.12  & -0.37 & 0.49 \\
Weak-white   & -0.03 & -0.82 & 0.79 & -0.10 & -0.59 & 0.49 \\
Strong-black & -0.05 & -0.32 & 0.27 & -0.12 & -0.25 & 0.13 \\
Human        & 0.02  & -0.12 & 0.14 & 0.03  & -0.04 & 0.07 \\
\hline
\end{tabular}

\caption{Human and AI EGS on XSum: high- vs. low-entropy regions.
High-20/Low-20 = mean EGS of top/bottom 20\% of tokens by entropy; High-50/Low-50 are defined similarly. Diff = EGS (Low)--EGS (High).
Settings follow Section~\ref{sec:entropy_calibration}.
}
\label{tab:direction_analysis}
\end{table}
\noindent\textbf{Trade-off between distribution shift and variance.} 
Top-$k$ filtering involves a trade-off between two effects.
Removing low-entropy tokens may push the AI distribution away from the human distribution (beneficial), but filtering also reduces token count and increases EGS variance, flattening the distribution (harmful). 
Figure~\ref{fig:variance_ratio} shows EGS variance increasing with the filtering ratio across three strategies.
Therefore, the net benefit of filtering depends on whether the separation gain outweighs the variance cost.
A toy Gaussian view in Appendix~\ref{app:gaussian_toy} further shows how mean shifts and variance changes jointly determine AUROC.
Figure~\ref{fig:filtering_histogram} compares AI and human distributions before and after top-$k$ filtering across the three settings.



We now explain why top-$k$ filtering exhibits distinct behaviors across the three settings.

\noindent\textbf{Weak-black: filtering helps.}
As discussed in Section~\ref{sec:entropy_calibration}, weak source LMs exhibit weak $L$--$E$ coupling ($\beta=0.809$) and produce more entropy-miscalibrated tokens. Figure~\ref{fig:bucket-heatmap} (see Appendix~\ref{app:heatmap_bucket}) shows EGS contributions by entropy buckets (high$\to$low) on XSum, with brighter cells indicating larger EGS. These high-EGS tokens invade the human region and harm detection. In the Weak-black setting, high-EGS tokens concentrate in low-entropy buckets, forming a large harmful region. Removing them shifts AI-side distribution downward, improving human-AI separation and boosting detection. 

Although filtering increases EGS variance, the separation gain dominates until the ratio reaches $\sim$50\%, after which performance gradually declines. This explains the inverted U-shaped pattern observed in Section~\ref{sec:filtering_ratio}.

\noindent\textbf{Weak-white: filtering neutral.}
In this setting, weak source LMs show weak $L$--$E$ coupling ($\beta=0.807$) and entropy miscalibration, as in Weak-black. Figure~\ref{fig:bucket-heatmap} confirms the same pattern: high EGS concentrates in low-entropy buckets, forming a harmful region.

However, because the surrogate matches the source model, distinguishing human from AI texts becomes easier~\citep{bao2025glimpse}. Human and AI distributions are well-separated with little overlap (see  Figure~\ref{fig:filtering_histogram}). Thus, zero-shot detectors generally perform better in white-box than black-box settings~\citep{mitchell2023detectgpt,baofast}.

Therefore, while top-$k$ filtering removes harmful low-entropy tokens and further reduces overlap, the benefit is offset by the variance cost of token removal. This explains the plateau up to a ratio of about 0.6, followed by a monotonic decline.

\noindent\textbf{Strong-black: filtering hurts.}
Unlike weak source LMs, the Strong-black setting shows stronger $L$--$E$ coupling ($\beta=0.943$) and better entropy calibration, leading to a small EGS gap between low- and high-entropy tokens (see Table~\ref{tab:direction_analysis}). Accordingly, the heatmap (Figure~\ref{fig:bucket-heatmap}) shows no large, continuous low-entropy region with high EGS. Thus, in Strong-black, low-entropy tokens do not form a reliable harmful region and provide weak detection signals.

Removing low-entropy tokens brings little benefit and cannot compensate for the variance cost introduced by token removal. Consequently, top-$k$ filtering performance starts declining early ($\sim$20\%) and continues to drop without an initial plateau.

\subsection{Controlled Validation of the Mechanism}
\label{sec:controlled_source}
We further validate the weak--strong difference on additional source LMs under a controlled generation protocol.

\begin{table}[t]
\centering
\footnotesize
\setlength{\tabcolsep}{2.5pt}
\renewcommand{\arraystretch}{1.05}
\begin{tabular}{@{}lcccc@{}}
    \hline
    \textbf{Source LM}
    & $\boldsymbol{\beta}$
    & \textbf{EGS (0)}
    & \textbf{EGS (50\%)}
    & $\boldsymbol{\Delta}$ \\
    \hline
    Claude-4
    & 0.9565 & 0.6405 & 0.5815 & $-$0.0589 \\
    DeepSeek-V4
    & 0.9190 & 0.7389 & 0.7115 & $-$0.0273 \\
    Gemini-2.5
    & 0.9324 & 0.8173 & 0.7433 & $-$0.0740 \\
    \hline
    Falcon-7B-Instruct
    & 0.8092 & 0.8306 & 0.8907 & $+$0.0601 \\
    \hline
\end{tabular}
\caption{
Controlled validation across additional source models (surrogate: GPT-2-XL).
$\Delta$ denotes the AUROC difference between EGS (50\%) and EGS (0).
All results are averaged across the three datasets.
}
\label{tab:controlled_source}
\end{table}

As shown in Table~\ref{tab:controlled_source}, all three strong source models exhibit high $L$--$E$ coupling and consistent performance degradation after filtering.
In contrast, Falcon-7B-Instruct retains substantially weaker coupling and benefits from filtering.
Notably, DeepSeek-V4 and Falcon-7B-Instruct are both open-weight instruction-tuned models, indicating that openness or instruction tuning alone does not explain the observed weak--strong difference.

To independently validate the calibration interpretation with an external measure, we compute token-level ECE \citep{guo2017calibration} based solely on the mismatch between the surrogate's top-1 confidence and top-1 correctness. Weak-source settings exhibit higher AI-side ECE than strong-source settings (0.0238 vs. 0.0145). 
The detailed formula is shown in Appendix~\ref{app:ece}.
Across all 27 settings, AI-side ECE strongly correlates with the AUROC gain from filtering (Pearson $r=0.7719$; Spearman $\rho=0.7900$).
These results provide independent evidence that filtering success is not an EGS-specific pattern.

\section{Conclusion}
\label{sec:conclusion}


In this work, we challenge the prevailing assumption that more token evidence is always better. Using top-$k$ filtering as a probe, we find that removing low-entropy tokens substantially improves detection in weak-source settings by shifting the AI-side distribution away from human text, but fails in strong-source settings. We attribute this discrepancy to entropy calibration: harmful low-entropy contributions emerge in the weak-source settings, but not in the strong-source settings. We further reveal that filtering entails a two-sided trade-off between distribution shift and increased variance, offering new insights into discriminative evidence in AIGT detection.

\section*{Limitations}



\noindent(1) This work analyzes the Entropy Gap Score (EGS) for single‑surrogate detection and introduces an EGS‑based filtering strategy as a diagnostic probe. Developing adaptive strategies and extending the analysis to multi-surrogate scores (e.g., Binoculars~\citep{hans2024spotting}) remain important directions for future work.

\noindent(2) Our evaluation focuses on several widely used English benchmarks and selected rewriting attacks. Broader robustness evaluations in more realistic settings, including multilingual text and additional distribution shifts, remain to be explored.

\noindent(3) The upstream factors that determine the entropy-calibration regimes and filtering behavior are not yet fully understood. Future work could investigate how training data, model scale, and decoding strategies jointly shape $L$--$E$ coupling and entropy calibration.

\section*{Ethical Considerations}
This work analyzes AI-generated text detection in controlled benchmark settings and is not intended as a deployed detection system. 
We use publicly available corpora and benchmarks under their licenses and do not collect private user data. 
Our results further show that detector behavior depends on source models, surrogate models, and token-level calibration, suggesting that detection scores should not be used as standalone evidence for punitive decisions.

\bibliography{custom}

\appendix
\begin{table*}
  \centering
  \begin{tabular}{ccc}
    \hline
    \textbf{Model} & \textbf{Model File/Service} & \textbf{Category} \\
    \hline
    GPT-2-XL~\citep{radford2019language}        & openai-community/gpt2-xl  & Weak   \\
    OPT-2.7B~\citep{zhang2022opt}               & facebook/opt-2.7b         & Weak   \\
    GPT-Neo-2.7B~\citep{black2021gpt}           & EleutherAI/gpt-neo-2.7B   & Weak   \\
    GPT-J-6B~\citep{wang2021gpt}                & EleutherAI/gpt-j-6B       & Weak   \\
    Falcon-7B~\citep{penedo2023refinedweb}      & tiiuae/falcon-7b          & Weak   \\
    Falcon-7B-Instruct~\citep{penedo2023refinedweb}   & tiiuae/falcon-7b-instruct          & Weak   \\
    Falcon-40B~\citep{penedo2023refinedweb}     & tiiuae/falcon-40b               & Weak   \\
    BLOOM-7.1B~\citep{workshop2022bloom}        & bigscience/bloom-7b1      & Weak   \\
    Qwen2.5-3B~\citep{yang2025qwen25}           & Qwen/Qwen2.5-3B           & Weak       \\
    GPT-4-Turbo~\citep{achiam2023gpt}           & OpenAI                    & Strong \\
    GPT-4o~\citep{hurst2024gpt}                 & OpenAI                    & Strong \\
    Claude-4~\citep{anthropic2025claude4}       & Anthropic                 & Strong \\
    DeepSeek-V4~\citep{xu2026deepseek}          & DeepSeek                  & Strong \\
    Gemini-2.5~\citep{comanici2025gemini}       & Google                    & Strong \\
    \hline
  \end{tabular}
  
  \caption{Details of the source and surrogate models.}
  \label{tab:source_models}
\end{table*}

\begin{table*}[t]

\centering

\begin{tabular}{lrrrrrr}
\hline
Setting & $k=2$ & $k=5$ & $k=10$ & $k=20$ & $k=50$ & $k=100$\\
\hline
Weak-black & +0.0671 & +0.0746 & +0.0808 & +0.0825 & +0.0826 & +0.0834\\
Strong-black & -0.0391 & -0.0400 & -0.0330 & -0.02917 & -0.0256 & -0.0276\\

\hline
\end{tabular}
\caption{Sensitivity to the choice of $k$ under the GPT-2-XL surrogate. We report the average $\Delta$AUROC of EGS (50\%) relative to EGS (0\%) across datasets and source LMs.}
\label{tab:k_sensitivity}
\end{table*}

\begin{table*}[t]
  \centering
  \small
  \setlength{\tabcolsep}{3pt}
  \renewcommand{\arraystretch}{1.18}
  \begin{tabular}{l*{12}{c}}
    \hline
    \textbf{Methods}
    & \multicolumn{2}{c}{\textbf{Falcon-7B}}
    & \multicolumn{2}{c}{\textbf{GPT-J}}
    & \multicolumn{2}{c}{\textbf{OPT-2.7B}}
    & \multicolumn{2}{c}{\textbf{GPT-Neo-2.7B}}
    & \multicolumn{2}{c}{\textbf{BLOOM-7.1B}}
    & \multicolumn{2}{c}{\textbf{Avg.}} \\
    \cline{2-13}

    & \textbf{Self} & \textbf{GPT2}
    & \textbf{Self} & \textbf{GPT2}
    & \textbf{Self} & \textbf{GPT2}
    & \textbf{Self} & \textbf{GPT2}
    & \textbf{Self} & \textbf{GPT2}
    & \textbf{Self} & \textbf{GPT2} \\
    
    \hline
    \multicolumn{13}{c}{\textbf{XSum}} \\
    \hline
    
    Log-Likelihood    & 0.6856 & 0.4764 & 0.8146 & 0.5983 & 0.8146 & 0.6349 & 0.8668 & 0.5668 & 0.8529 & 0.4402 & 0.8069 & 0.5433 \\
    Entropy           & 0.3292 & 0.3834 & 0.4019 & 0.3866 & 0.4567 & 0.4381 & 0.4018 & 0.3392 & 0.3043 & 0.2941 & 0.3788 & 0.3683 \\
    LogRank           & 0.7364 & 0.5546 & 0.8498 & 0.6565 & 0.8464 & 0.6930 & 0.9062 & 0.6337 & 0.9072 & 0.5398 & 0.8492 & 0.6155 \\
    DetectGPT         & 0.8207 & 0.5292 & 0.9023 & 0.5988 & 0.9036 & 0.6147 & 0.9563 & 0.6517 & 0.9814 & 0.5079 & 0.9129 & 0.5805 \\
    Fast-DetectGPT    & 0.9642 & 0.6179 & 0.9811 & 0.7790 & 0.9698 & 0.7746 & 0.9901 & 0.8016 & 0.9920 & 0.7138 & 0.9794 & 0.7374 \\
    Lastde++          & \textbf{0.9844} & \underline{0.6657} & \textbf{0.9910} & \underline{0.8494} & \textbf{0.9845} & \underline{0.8367} & \textbf{0.9980} & \underline{0.8840} & \textbf{0.9992} & \underline{0.7753} & \textbf{0.9914} & \underline{0.8022} \\
    SpecDetect++      & \underline{0.9788} & 0.6159 & \underline{0.9821} & 0.8324 & \underline{0.9806} & 0.7953 & \underline{0.9930} & 0.8612 & \underline{0.9940} & 0.7420 & \underline{0.9857} & 0.7694 \\
    EGS (0\%)          & 0.9635 & 0.6223 & 0.9785 & 0.7813 & 0.9669 & 0.7762 & 0.9880 & 0.8039 & 0.9930 & 0.7142 & 0.9780 & 0.7396 \\
    EGS (50\%)         & 0.9677 & \textbf{0.8414} & 0.9733 & \textbf{0.8959} & 0.9605 & \textbf{0.8764} & 0.9874 & \textbf{0.9131} & 0.9902 & \textbf{0.8978} & 0.9758 & \textbf{0.8849} \\
    \hline
    EGS (opt)      & 0.9677 & 0.8414 & 0.9808 & 0.8959 & 0.9686 & 0.8764 & 0.9899 & 0.9131 & 0.9944 & 0.8978 & 0.9803 & 0.8849 \\
    optimal~$\theta$  & 0.5 & 0.5 & 0.3 & 0.5 & 0.3 & 0.5 & 0.2 & 0.5 & 0.2 & 0.5 & 0.3 & 0.5 \\
    \hline
    \multicolumn{13}{c}{\textbf{Reddit}} \\
    \hline
    
    Log-Likelihood    & 0.8348 & 0.7210 & 0.8684 & 0.7998 & 0.9140 & 0.7198 & 0.9151 & 0.8120 & 0.9089 & 0.7708 & 0.8882 & 0.7647 \\
    Entropy           & 0.4969 & 0.5377 & 0.5183 & 0.5428 & 0.4759 & 0.4528 & 0.5397 & 0.5094 & 0.4673 & 0.4914 & 0.4996 & 0.5068 \\
    LogRank           & 0.8698 & 0.7647 & 0.9012 & 0.8472 & 0.9384 & 0.7824 & 0.9420 & 0.8540 & 0.9382 & 0.8267 & 0.9179 & 0.8150 \\
    DetectGPT         & 0.8792 & 0.6932 & 0.9496 & 0.7875 & 0.9461 & 0.8388 & 0.9815 & 0.8852 & 0.9632 & 0.7654 & 0.9439 & 0.7940 \\
    Fast-DetectGPT    & 0.9759 & 0.8178 & 0.9859 & 0.9095 & 0.9904 & 0.9072 & 0.9965 & 0.9402 & \underline{0.9968} & 0.9125 & 0.9891 & 0.8974 \\
    Lastde++          & \textbf{0.9872} & \underline{0.8592} & \textbf{0.9930} & \underline{0.9494} & \textbf{0.9971} & \underline{0.9527} & \textbf{0.9990} & \underline{0.9699} & \textbf{0.9994} & \underline{0.9402} & \textbf{0.9951} & \underline{0.9343} \\
    SpecDetect++      & 0.9817 & 0.8262 & \underline{0.9919} & 0.9219 & 0.9913 & 0.9294 & 0.9924 & 0.9463 & 0.9955 & 0.9049 & \underline{0.9906} & 0.9057 \\
    EGS (0\%)          & 0.9761 & 0.8167 & 0.9836 & 0.9093 & \underline{0.9915} & 0.9079 & 0.9953 & 0.9406 & 0.9965 & 0.9127 & 0.9886 & 0.8974 \\
    EGS (50\%)         & \underline{0.9818} & \textbf{0.9061} & 0.9891 & \textbf{0.9531} & 0.9912 & \textbf{0.9537} & \underline{0.9966} & \textbf{0.9707} & 0.9940 & \textbf{0.9505} & 0.9905 & \textbf{0.9468} \\
    \hline
    EGS (opt)      & 0.9830 & 0.9061 & 0.9904 & 0.9531 & 0.9928 & 0.9555 & 0.9966 & 0.9707 & 0.9981 & 0.9555 & 0.9922 & 0.9482 \\
    optimal~$\theta$  & 0.6 & 0.5 & 0.7 & 0.5 & 0.3 & 0.6 & 0.5 & 0.5 & 0.2 & 0.3 & 0.46 & 0.48 \\
    \hline
    \multicolumn{13}{c}{\textbf{WritingPrompts}} \\
    \hline
    
    Log-Likelihood    & 0.8616 & 0.7657 & 0.9293 & 0.8730 & 0.9397 & 0.8181 & 0.9477 & 0.8828 & 0.9365 & 0.8174 & 0.9230 & 0.8314 \\
    Entropy           & 0.5225 & 0.5538 & 0.5802 & 0.6168 & 0.5582 & 0.5388 & 0.6256 & 0.5807 & 0.4893 & 0.5186 & 0.5552 & 0.5618 \\
    LogRank           & 0.8918 & 0.8111 & 0.9528 & 0.9067 & 0.9594 & 0.8752 & 0.9653 & 0.9196 & 0.9612 & 0.8786 & 0.9461 & 0.8782 \\
    DetectGPT         & 0.8495 & 0.6550 & 0.9425 & 0.8105 & 0.9584 & 0.8429 & 0.9701 & 0.8432 & 0.9696 & 0.7388 & 0.9380 & 0.7781 \\
    Fast-DetectGPT    & 0.9852 & 0.8538 & 0.9921 & 0.9339 & 0.9968 & 0.9371 & 0.9980 & 0.9553 & 0.9980 & 0.9257 & 0.9940 & 0.9212 \\
    Lastde++          & \textbf{0.9952} & \underline{0.9116} & \textbf{0.9980} & \textbf{0.9664} & \textbf{0.9996} & \underline{0.9719} & \textbf{1.0000} & \underline{0.9791} & \textbf{1.0000} & \underline{0.9704} & \textbf{0.9986} & \underline{0.9599} \\
    SpecDetect++      & \underline{0.9912} & 0.9032 & 0.9941 & 0.9535 & \underline{0.9974} & 0.9704 & 0.9981 & 0.9725 & \underline{0.9989} & 0.9661 & \underline{0.9959} & 0.9531 \\
    EGS (0\%)          & 0.9881 & 0.8550 & 0.9915 & 0.9332 & 0.9961 & 0.9345 & 0.9971 & 0.9536 & 0.9980 & 0.9265 & 0.9942 & 0.9206 \\
    EGS (50\%)         & 0.9911 & \textbf{0.9360} & \underline{0.9948} & \underline{0.9626} & 0.9940 & \textbf{0.9793} & \underline{0.9987} & \textbf{0.9879} & 0.9968 & \textbf{0.9755} & 0.9951 & \textbf{0.9683} \\
    
    \hline
    EGS (opt)      & 0.9913 & 0.9370 & 0.9952 & 0.9644 & 0.9968 & 0.9793 & 0.9995 & 0.9900 & 0.9987 & 0.9780 & 0.9963 & 0.9698 \\
    optimal~$\theta$  & 0.6 & 0.6 & 0.4 & 0.6 & 0.3 & 0.5 & 0.4 & 0.6 & 0.2 & 0.6 & 0.38 & 0.58 \\
    \hline
  \end{tabular}

  \caption{Detection results for text generated by five weak source LMs. For each source model, ``Self'' denotes the white-box setting, while ``GPT2'' denotes the fixed-surrogate setting where GPT-2-XL is used as the surrogate model. EGS (opt) represents the optimal ratio performance while EGS ($\theta$) represents performance under a fixed ratio $\theta$. The Avg. columns report the mean AUROC over the five weak source LMs under the corresponding setting.}
  \label{tab:weak_results_detail}
\end{table*}

\section{Derivation of the Entropy Gap Score and Its Relation to Fast-DetectGPT}
\label{app:derivation}
To clarify the connection between our Entropy Gap Score and the conditional probability curvature used in Fast-DetectGPT~\citep{baofast}, we rewrite both scores under the same token-level notation.

Given an input passage $x=(x_1,\ldots,x_N)$ and a scoring model $M$, the token-level self-information is
\begin{equation}
L_i = -\log p_M(x_i \mid x_{<i}).
\end{equation}
The expected self-information under the model's next-token distribution is
\begin{equation}
\begin{aligned}
E_i
&= \mathbb{E}_{v\sim p_M(\cdot \mid x_{<i})}
\left[-\log p_M(v \mid x_{<i})\right] \\
&= -\sum_{v\in\mathcal{V}}
p_M(v\mid x_{<i})
\log p_M(v\mid x_{<i}),
\end{aligned}
\end{equation}
where $\mathcal{V}$ denotes the vocabulary. 
Our Entropy Gap Score is defined as the token-average gap between the observed self-information and its model-expected value:
\begin{equation}
\mathrm{EGS}(x)
=
\frac{1}{N}\sum_{i=1}^{N}(L_i-E_i).
\end{equation}
Fast-DetectGPT defines its conditional probability curvature as
\begin{equation}
d(x,p_\theta,q_\phi)
=
\frac{\log p_\theta(x)-\tilde{\mu}}{\tilde{\sigma}},
\end{equation}
where
\begin{equation}
\tilde{\mu}
=
\mathbb{E}_{\tilde{x}\sim q_\phi(\tilde{x}\mid x)}
\left[\log p_\theta(\tilde{x}\mid x)\right].
\end{equation}
Here, $\log p_\theta(x)-\tilde{\mu}$ is the unnormalized curvature numerator, while $\tilde{\sigma}$ is used only for standardization.

We consider the special case where the perturbation distribution and the scoring model are the same, i.e., $q_\phi=p_\theta=p_M$. Under the token-wise decomposition, the expectation term becomes
\begin{equation}
\begin{aligned}
\tilde{\mu}_M
&=
\sum_{i=1}^{N}
\sum_{v\in\mathcal{V}}
p_M(v\mid x_{<i})
\log p_M(v\mid x_{<i}) \\
&=
-\sum_{i=1}^{N} E_i.
\end{aligned}
\end{equation}
\begin{equation}
d(x,p_M,q_M)
=
\frac{\log p_M(x)-\tilde{\mu}_M}{\tilde{\sigma}_M},
\end{equation}
Meanwhile, the log-likelihood of the observed passage is
\begin{equation}
\begin{aligned}
\log p_M(x)
&=
\sum_{i=1}^{N}
\log p_M(x_i\mid x_{<i}) \\
&=
-\sum_{i=1}^{N} L_i.
\end{aligned}
\end{equation}
Therefore, the unnormalized Fast-DetectGPT curvature numerator can be rewritten as
\begin{equation}
\begin{aligned}
\log p_M(x)-\tilde{\mu}_M
&=
-\sum_{i=1}^{N}L_i
+
\sum_{i=1}^{N}E_i \\
&=
-\sum_{i=1}^{N}(L_i-E_i) \\
&=
-N\cdot \mathrm{EGS}(x).
\end{aligned}
\end{equation}
This yields
\begin{equation}
\mathrm{EGS}(x)
=
-\frac{1}{N}
\left(
\log p_M(x)-\tilde{\mu}_M
\right).
\end{equation}
This shows that EGS corresponds to the token-averaged, unstandardized counterpart of the Fast-DetectGPT curvature numerator under $q_\phi=p_\theta=p_M$, up to a sign difference caused by using self-information rather than log-likelihood. 
Unlike Fast-DetectGPT, EGS does not divide by the standard deviation term $\tilde{\sigma}$ and therefore avoids the standardization step.

\section{Implementation Details}
\label{app:implementation-details}
This section describes the implementation details for our analysis experiments.
All experiments were performed on a Linux server with 8 NVIDIA 4090 24G GPUs.

We use XSum, WritingPrompts, and Reddit for detection.
Following~\citet{xutraining}, we filter out samples with text length less than 150 words, and randomly sample 150 of these as human-written examples for each dataset.
We then use the first 30 tokens as a prompt to generate a continuation from the source model.
To ensure fairness, we truncate all texts to 256 tokens before calculating the results and baselines.
For weak source LMs, we follow the default sampling configuration.

Table~\ref{tab:source_models} summarizes the source and surrogate models used in our experiments. 
For weak LMs, we use the corresponding open-source checkpoints listed in the table.
For GPT-4-Turbo and GPT-4o, we adopt the generated samples from \citet{xutraining}. For Claude-4, DeepSeek-V4, and Gemini-2.5, we generate continuations following the same data construction and prompting protocol.

\section{Additional Results}
\label{app:additional_results}
\subsection{Weak Source Details}
\label{app:weak_details}
Table~\ref{tab:weak_results_detail} reports the detailed weak source detection results across all datasets and source--surrogate settings. The results support the main finding that top-$k$ filtering consistently improves performance under weak-source black-box detection.

\subsection{Heatmap of EGS Bucket}
\label{app:heatmap_bucket}
\begin{figure}[t]
\includegraphics[width=\linewidth]{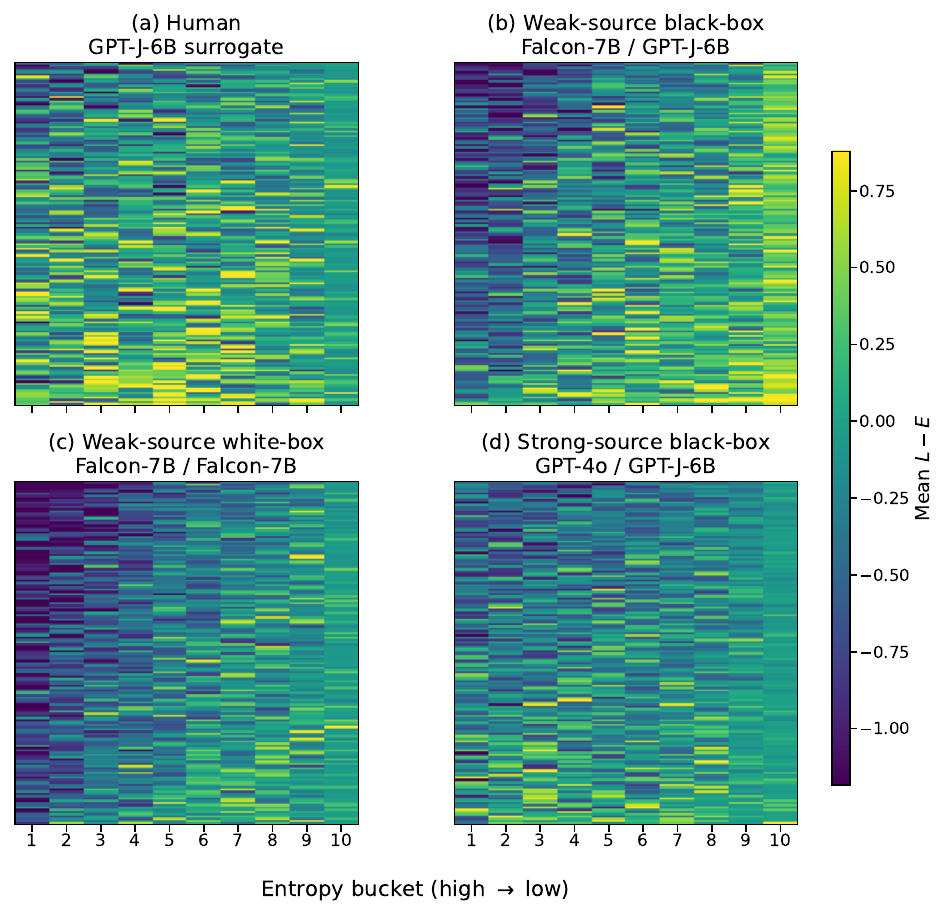}
\caption{
EGS contributions across entropy buckets on XSum. Panels: (a) Human, (b) Weak-black, (c) Weak-white, (d) Strong-black (see Section~\ref{sec:entropy_calibration}). Each row: a text with tokens sorted high to low entropy. Columns: 150 stacked texts. Mean EGS per entropy bucket; brightness = EGS magnitude (brighter = larger).
}
\label{fig:bucket-heatmap}
\end{figure}

Figure~\ref{fig:bucket-heatmap} shows EGS contributions across entropy buckets on XSum across four settings (Human, Weak-black, Weak-white, Strong-black; see Section~\ref{sec:entropy_calibration}). Each row represents a text with tokens sorted from high to low entropy, and columns stack 150 texts. Mean EGS is computed per entropy bucket, with brightness indicating magnitude (brighter = larger EGS).

\subsection{Token Filtering Overlap}
\label{app:cross_surrogate_filtering}

Table~\ref{tab:cross_surrogate_jaccard} shows the overlap of token regions selected by different surrogates and tokenizers at removal ratios of 0.2 and 0.5. The regions selected by top-$k$ filtering exhibit substantially higher overlap than random selection, indicating that the filtered tokens are not specific to a single surrogate or tokenizer.

\subsection{Held-out Ratio Selection}
\label{app:ratio_selection}

To examine whether the fixed filtering ratio can be selected without test-set tuning, we use GPT-2-XL as the surrogate model and conduct leave-one-domain-out (LODO) and leave-one-source-out (LOSO) evaluation over the five weak source LMs. 

For each held-out domain or source LM, we select a single filtering ratio from $\{0,0.1,\ldots,0.9\}$ based on the average AUROC over the remaining folds, and then evaluate the selected ratio on the held-out fold.
The selected ratio is 0.5 in all three LODO folds and all five LOSO folds, supporting the fixed 50\% filtering ratio used in the main experiments.

\subsection{Summary across All Pairs}
\label{summary_all_pairs}
Table~\ref{tab:beta_summary} summarizes the linear regression slope $\beta$, fixed-ratio filtering gains, and optimal filtering ratios across all datasets averaged by source--surrogate settings. The results indicate the phenomenon is consistent and $\Delta$AUROC@50\% is correlated with $\beta$.

\begin{table*}[t]
\centering
\begin{tabular}{lccc}
\hline
Setting & $\Delta$AUROC & 95\% CI & \# Pairs \\
\hline
Weak-black & 0.058 & [0.055, 0.061] & 90 \\
Weak-white & 0.000 & [-0.001, 0.002] & 18 \\
Strong-black & -0.044 & [-0.048, -0.039] & 36 \\
\hline
\end{tabular}
\caption{
Paired bootstrap confidence intervals for the AUROC change of fixed top-$k$ filtering at $\theta=0.5$ relative to unfiltered EGS.
For each bootstrap iteration, we resample text pairs with replacement within each dataset--source--surrogate pair, compute $\Delta$AUROC, and then average $\Delta$AUROC over all pairs in  the same setting. 
We use 2,000 bootstrap iterations.
This analysis uses the six main surrogate models in our filtering experiments.
}
\label{tab:bootstrap_ci}
\end{table*}

\begin{table*}[t]
\centering
\begin{tabular}{lrrrr}
\hline
Setting & \# Pairs & EGS (0) AUROC & EGS (50\%) AUROC & $\Delta$AUROC \\
\hline
Writing / Falcon-7B & 150   & $0.8788 \pm 0.0207$ & $0.9551 \pm 0.0106$ & $+0.0763 \pm 0.0123$ \\
Writing / Falcon-7B & 500   & $0.8647 \pm 0.0062$ & $0.9471 \pm 0.0021$ & $+0.0824 \pm 0.0059$ \\
Writing / Falcon-7B & 1,000 & 0.8621              & 0.9463              & +0.0842              \\
\hline
XSum / GPT-J-6B     & 150   & $0.7419 \pm 0.0356$ & $0.8888 \pm 0.0223$ & $+0.1469 \pm 0.0267$ \\
XSum / GPT-J-6B     & 500   & $0.7535 \pm 0.0116$ & $0.8870 \pm 0.0064$ & $+0.1336 \pm 0.0068$ \\
XSum / GPT-J-6B     & 1,000 & 0.7585              & 0.8886              & +0.1301              \\
\hline
\end{tabular}
\caption{Detection AUROC results with different numbers of paired samples. Results for 150 and 500 pairs are reported as mean $\pm$ standard deviation over five random subsets drawn from the 1,000-pair pool. EGS (0) denotes the unfiltered EGS, while EGS (50\%) removes 50\% of tokens using top-$k$ filtering.}
\label{tab:sample_size}

\end{table*}

\subsection{Token-Level Calibration Analysis}
\label{app:ece}

To provide an external calibration measure independent of the EGS
formulation, we compute token-level expected calibration error (ECE)
using only the surrogate model's top-1 confidence and correctness.
For each token position, the confidence is the probability assigned to
the surrogate's top-1 prediction, while correctness indicates whether
the observed token matches this prediction.
We partition confidence values into $M=15$ equal-width bins and compute

\begin{equation}
\mathrm{ECE}
=
\sum_{m=1}^{M}
\frac{|B_m|}{N}
\left|
\operatorname{acc}(B_m)
-
\operatorname{conf}(B_m)
\right|,
\end{equation}

where $B_m$ denotes the set of token positions in the $m$-th confidence
bin, $N$ is the total number of token positions,
$\operatorname{acc}(B_m)$ is the fraction of correct top-1 predictions
in the bin, and $\operatorname{conf}(B_m)$ is their average top-1
confidence.

We compute ECE separately for human-written and AI-generated tokens and
use the AI-side ECE for the analysis in the main text.
Weak-source settings exhibit higher AI-side ECE than strong-source
settings (0.0238 vs.\ 0.0145).
To examine its relationship with filtering, we define the filtering gain
as the AUROC difference between 50\% filtering and no filtering.
Across all 27 settings, AI-side ECE strongly correlates with filtering
gain (Pearson $r=0.7719$; Spearman $\rho=0.7900$), providing calibration
evidence independent of the EGS-based analysis.

\subsection{A Toy Gaussian View of the Shift--Variance Tradeoff}
\label{app:gaussian_toy}
To further illustrate the shift--variance tradeoff, suppose that the document-level EGS scores of HWT and AIGT follow Gaussian distributions,
$S_H\sim\mathcal{N}(\mu_H,\sigma_H^2)$ and
$S_A\sim\mathcal{N}(\mu_A,\sigma_A^2)$. 
When smaller scores indicate AIGT, the AUROC can be written as
\[
\mathrm{AUROC}
=
P(S_H>S_A)
=
\Phi\left(
\frac{\mu_H-\mu_A}{\sqrt{\sigma_H^2+\sigma_A^2}}
\right),
\]
where $\Phi$ is the standard normal cumulative distribution function.
This toy model shows that AUROC depends not only on the mean separation $\mu_H-\mu_A$, but also on the two-sided variance cost $\sigma_A^2+\sigma_H^2$.

Empirically, as shown in Table~\ref{tab:gaussian_auc}, this Gaussian approximation closely matches the observed AUROC, with an average gap below 0.004 across the main settings.

\begin{table*}[t]
\centering
\begin{tabular}{llccc}
\hline
Setting & $\theta$ & Emp. AUROC & Gaussian AUROC & Gap \\
\hline
Strong-black & 0.0 & 0.882 & 0.880 & 0.002 \\
Strong-black & 0.5 & 0.838 & 0.838 & 0.001 \\
Weak-black & 0.0 & 0.865 & 0.861 & 0.004 \\
Weak-black & 0.5 & 0.923 & 0.921 & 0.003 \\
Weak-white & 0.0 & 0.988 & 0.987 & 0.002 \\
Weak-white & 0.5 & 0.989 & 0.987 & 0.001 \\
\hline
\end{tabular}
\caption{
Empirical AUROC and the Gaussian approximation under the toy
shift--variance model. $\theta$: filtering ratio. The small gaps show that the model captures the main effect of mean separation and variance on AUROC.
}
\label{tab:gaussian_auc}
\end{table*}

\begin{table*}[t]
\centering
\begin{tabular}{llccc}
\hline
Surrogate Pair & Alignment Level & Remove Ratio & Observed Jaccard & Random Jaccard \\
\hline
GPT-2-XL / GPT-J-6B  & Token index    & 0.2 & 0.7580 & 0.1130 \\
GPT-2-XL / GPT-J-6B  & Token index    & 0.5 & 0.8258 & 0.3362 \\
GPT-2-XL / Falcon-7B & Character span & 0.2 & 0.6228 & 0.1131 \\
GPT-2-XL / Falcon-7B & Character span & 0.5 & 0.7345 & 0.3356 \\
\hline
\end{tabular}
\caption{Cross-surrogate consistency of selected token regions. For surrogates sharing the same tokenizer, overlap is computed over token indices; for different tokenizers, selected tokens are mapped to character spans before computing Jaccard overlap.}
\label{tab:cross_surrogate_jaccard}
\end{table*}

\begin{table*}[t]
\centering
\begin{tabular}{llcccc}
\hline
Dataset & Setting & \#Pairs & $\beta$ & $\Delta$AUROC@50\% & optimal~$\theta$ \\
\hline
XSum & Weak-white   & 6  & $0.7891 \pm 0.0181$ & $-0.0016 \pm 0.0040$ & $0.3000$ \\
XSum & Weak-black   & 30 & $0.8162 \pm 0.0324$ & $0.1021 \pm 0.0462$ & $0.5267$ \\
XSum & Strong-black & 12 & $0.9253 \pm 0.0152$ & $-0.0440 \pm 0.0122$ & $0.1333$ \\
\hline
WritingPrompts & Weak-white   & 6  & $0.7720 \pm 0.0149$ & $0.0010 \pm 0.0022$ & $0.4000$ \\
WritingPrompts & Weak-black   & 30 & $0.7909 \pm 0.0286$ & $0.0322 \pm 0.0162$ & $0.5100$ \\
WritingPrompts & Strong-black & 12 & $0.9469 \pm 0.0148$ & $-0.0528 \pm 0.0157$ & $0.0250$ \\
\hline
Reddit & Weak-white   & 6  & $0.7809 \pm 0.0184$ & $0.0015 \pm 0.0034$ & $0.3833$ \\
Reddit & Weak-black   & 30 & $0.7983 \pm 0.0301$ & $0.0403 \pm 0.0201$ & $0.5067$ \\
Reddit & Strong-black & 12 & $0.9166 \pm 0.0110$ & $-0.0350 \pm 0.0063$ & $0.0000$ \\
\hline
\end{tabular}
\caption{
Summary of $L$--$E$ coupling and top-$k$ filtering results in three datasets and three settings (Weak-white, Weak-black and Strong-black).
Values are averaged over source--surrogate pairs within each setting, with standard deviations reported after $\pm$.
$\beta$ denotes the regression slope, $\Delta$AUROC@50\% denotes the AUROC change after filtering out 50\% from both human and AI texts.
optimal~$\theta$ denotes the average optimal filtering ratio.
}
\label{tab:beta_summary}
\end{table*}

\section{Robustness Analysis}
\label{app:robustness}

\subsection{Sample-size Robustness}
\label{app:sample_robustness}
To assess whether the observed filtering effects are sensitive to 150 samples, we first compute paired bootstrap confidence intervals for the fixed filtering ratio $\theta=0.5$. Table~\ref{tab:bootstrap_ci} shows that the weak-source black-box gains and strong-source black-box degradation are both stable under resampling.

We also expand sample size from 150 to 1,000, as shown in Table~\ref{tab:sample_size}.
These results show that 150 paired samples are sufficient for reliable evaluation in AI-generated text detection, with larger sample sizes yielding consistent AUROC estimates and relative performance trends.

\subsection{Sensitivity to the Choice of $k$}
\label{app:k_sensitivity}
We vary $k \in \{2,5,10,20,50,100\}$ and report the average AUROC change of EGS (50\%) relative to EGS (0\%) across datasets and source LMs (surrogate: GPT-2-XL). As shown in Table~\ref{tab:k_sensitivity}, varying $k$ changes the magnitude of the filtering effect but the tendency remains.
To balance performance and computational cost, we use $k=10$ as the default setting in our experiments.

\subsection{Rewriting Attack Robustness}
\label{app:rewriting}
We use T5 for paraphrasing \citep{raffel2020exploring} and the OPUS-MT en-de and de-en models for English--German--English back-translation \citep{tiedemann2020opus}. All results use GPT-2-XL as the surrogate model.

\begin{figure*}[t]
    \centering
    \includegraphics[width=\linewidth]{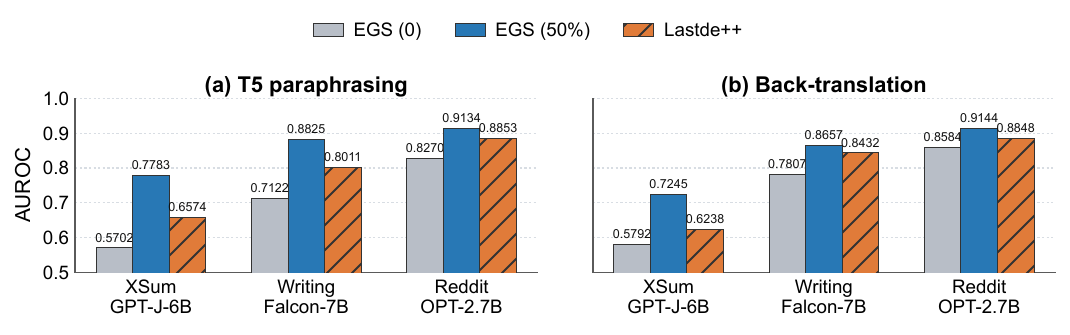}
    \caption{
Detection performance under two rewriting attacks (T5 paraphrasing and back-translation). GPT-2-XL is the surrogate, and AUROC values are shown above the bars.
}
\label{fig:attack_results}
\end{figure*}

As shown in Figure~\ref{fig:attack_results}, top-$k$ filtering at a fixed 50\% ratio consistently improves unfiltered EGS under both rewriting transformations and outperforms Lastde++ in all six settings.
The average AUROC gains over unfiltered EGS and Lastde++ are +0.1252 and +0.0639, respectively. These results demonstrate the robustness of top-$k$ filtering under rewriting attacks and show that the filtering phenomenon is not limited to raw AI-generated text.

\section{Qualitative Analysis of Filtered Tokens}
\label{app:qualitative_analysis}
We further inspect the tokens removed by top-$k$ filtering in a weak-source, black-box setting to examine whether they correspond to visibly anomalous text spans.
The following two examples are truncated to the first 100 tokens.
For each example, we apply top-$k$ filtering with a fixed 30\% filtering ratio and highlight these removed tokens.

\noindent\textbf{Example 1.}
Source: Falcon-7B; Surrogate: GPT-J-6B; Dataset: XSum.
Before filtering: EGS = -0.2263;
after filtering: EGS = -0.4969;
$\Delta$EGS = -0.2707.
\begin{quote}
 grant of £973,000 has been given\hl{ to} Wildcat Action - which involves more\hl{ than} 30 organisations, community groups\hl{ and} landowners. Over £100\hl{,}\hl{000}\hl{ of} the money\hl{ has}\hl{ been} allocated\hl{ to} the Wild\hl{cat} Task\hl{force}, an alliance\hl{ that} is being led\hl{ by} the charity.
\hl{The} project will run\hl{ the} programme between now\hl{ and} October 2018\hl{ on} a £650\hl{,}\hl{000} contract with the Department\hl{ for} Environment\hl{,}\hl{ Food}\hl{ and}\hl{ Rural}\hl{ Affairs}\hl{ (}\hl{Def}\hl{ra}\hl{).}
The plan will be to reintrodu\hl{ce} the mountain-l
\end{quote}

\noindent\textbf{Example 2.}
Source: GPT-2-XL; Surrogate: BLOOM-7.1B; Dataset: WritingPrompts.
Before filtering: EGS = 0.3830;
after filtering: EGS = -0.0350;
$\Delta$EGS = -0.4180.

\begin{quote}
 name is Death. You work\hl{ 168}\hl{ hours}\hl{ a}\hl{ week} as\hl{ the} re\hl{aper} of souls on planet Earth\hl{.} You're living on just a couple b\hl{ucks}\hl{ above}\hl{ the}\hl{ national}\hl{ poverty}\hl{ level}, you're eating in ram\hl{sh}\hl{ack}\hl{le} cem\hl{eter}\hl{ies} on the side\hl{-p}ay for a dead family\hl{'s} food stamp, and you keep a shot\hl{gun} in\hl{ your} car to scare the poor dev\hl{ils} away\hl{ from} your job.

You're not happy about it\hl{,} either\hl{.} You get fed\hl{ up} one\hl{ day}\hl{ while} watching a homeless man on\hl{ another} man's front
\end{quote}

Two examples show that the removed tokens are ordinary tokens. This indicates that the “harmful” tokens are not anomalous spans, but ordinary tokens whose probability-based evidence becomes unreliable under entropy miscalibration.
Removing these tokens leaves a subset that is more informative for detection.

\section{Discussion}
Our findings suggest that token-level filtering is not universally beneficial. 
Its effectiveness depends on whether the generated tokens are miscalibrated relative to human-written text. 
When this calibration gap is large, as in weak source LMs, low-entropy high-confidence tokens can introduce harmful EGS contributions, and filtering them yields consistent gains across surrogates.
When the source LM is better calibrated, the harmful low-entropy shift becomes weaker and less persistent, so filtering removes useful evidence and amplifies variance.

This observation also suggests a broader direction for AI-generated text detection. 
Rather than designing detectors only from aggregate likelihood statistics, future methods may explicitly exploit LM miscalibration, similar in spirit to out-of-distribution detection.
Moreover, the large gains obtained by filtering a small subset of harmful tokens indicate that token selection remains underexplored. 
\end{document}